\documentclass[a4paper,fleqn]{cas-sc}
\usepackage{subfig}
\usepackage{multirow}
\usepackage{amssymb}
\usepackage{pifont}
\usepackage{color}

\usepackage[numbers]{natbib}

\def\tsc#1{\csdef{#1}{\textsc{\lowercase{#1}}\xspace}}
\tsc{WGM}
\tsc{QE}

\begin{document}
\begin{sloppypar}
\let\WriteBookmarks\relax
\def\floatpagepagefraction{1}
\def\textpagefraction{.001}
\let\printorcid\relax

\shorttitle{Document Image Rectification Bases on Self-Adaptive Multitask Fusion}    

\shortauthors{Heng Li et~al.}  

\title [mode = title]{Document Image Rectification Bases on Self-Adaptive Multitask Fusion}  



%

\author[aff1]{Heng Li}



\ead{hengli.lh@outlook.com}


\credit{}

\affiliation[aff1]{organization={the School of Computer Science and Technology, Harbin Institute of Technology (Shenzhen)},
            city={Shenzhen},
            country={China}}

\affiliation[aff2]{organization={Peng Cheng Laboratory},
            city={Shenzhen},
            country={China}}

\author[aff1]{Xiangping Wu}
\cormark[1]

\ead{wxpleduole@gmail.com}


\credit{}


\cortext[1]{Corresponding author}



\author[aff1, aff2]{Qingcai Chen}
\ead{qingcai.chen@hit.edu.cn}

\begin{abstract}
Deformed document image rectification is essential for real-world document understanding tasks, such as layout analysis and text recognition. However, current multi-task methods---such as background removal, 3D coordinate prediction, and text line segmentation---often overlook the complementary features between tasks and their interactions. To address this gap, we propose a self-adaptive learnable multi-task fusion rectification network named \textit{SalmRec}. This network incorporates an inter-task feature aggregation module that adaptively improves the perception of geometric distortions, enhances feature complementarity, and reduces negative interference. We also introduce a gating mechanism to balance features both within global tasks and between local tasks effectively. Experimental results on two English benchmarks (DIR300 and DocUNet) and one Chinese benchmark (DocReal) demonstrate that our method significantly improves rectification performance. Ablation studies further highlight the positive impact of different tasks on dewarping and the effectiveness of our proposed module.
\end{abstract}




\maketitle

\section{Introduction}\label{}
In recent years, advancements in computer hardware and the growing availability of internet data have accelerated the development of AI fields like image vision \cite{Wang2024DeepLI, Xiao2023TableDF, Kirillov2023SegmentA, Xie2022C2AC, CHENG2025105392, ZHAI2025105399}, multitask learning \cite{Bi2022MultitaskFL, Wang2022DomainAM, Feng2023MultiTaskPR, HE2024105109, ZIJUN2024105158, DOU2023104719}, and geometric representation learning \cite{Zheng2020EnhancingGF, Lu2022ExcavationRL, Vilnis2021GeometricRL, Poklukar2022GeometricMC, YAN2025105423}. Meanwhile, the widespread use of smartphones and portable cameras has led to an increase in electronic images of documents, such as academic papers, posters, receipts, invoices, contracts, transaction orders, and street views.

The ability to automatically extract structured data---such as text, visual question answering, and document layout analysis \cite{Li2024VOLTERVC, Liu2024TextAdapterSD, Wei2024GeneralOT, Zhang2022MultimodalPB, Bi2023SRRVAN, Liao2024DocLayLLMAE, Zhang2024DocKylinAL, Feng2023DocPediaUT}---from these images can greatly enhance convenience in daily life. However, real-world images are often affected by human and environmental factors, leading to varied presentations across different scenes. For instance, when capturing images with mobile devices, factors like device positioning, lighting, and paper angle can cause varying degrees of distortion, making it difficult to extract useful information. To address this, some researchers have focused on flattening distorted document images to enhance readability and improve document understanding accuracy.

Significant progress has been made in distorted document image rectification. Some previous methods \cite{Ma2018DocUNetDI, Liu2020GeometricRO, Xie2022DocumentDW, Xue2022FourierDR} directly regress dense 2-Dimension (2D) deformation fields or sparse control points to predict the pixel position mapping between the input distorted image and the scanned image. No additional guidance information is used in these single task based works. But when the image is strongly disturbed by lighting, shadows, the model's perception of the document foreground boundary is often inaccurate.

With the release of the synthetic distorted document training dataset Doc3D \cite{Das2019DewarpNetSD}, subsequent works gradually guided the prediction of the two-dimensional deformation field through a multi-task pipeline. This includes the extraction of document foreground \cite{Feng2021DocTrDI,Zhang2022MariorMR,li2023foreground}, prediction of intermediate feature 3D coordinates in the distorted paper imaging process \cite{Das2019DewarpNetSD, Feng2022GeometricRL} and UV maps \cite{li2023layout}. These external features assist the network in segmenting the document's foreground, particularly its boundary. To achieve finer granularity, text lines \cite{Jiang2022RevisitingDI, Feng2022GeometricRL, li2023foreground} and layouts \cite{li2023layout} are also extracted, enhancing readability and improving local correction accuracy.

These methods have leveraged multi-task learning to explore the impact of various imaging features on rectification, achieving inspiring improvements in correcting distorted images. However, these approaches still face several challenges. First, previous methods often focus on combining global and local features, such as 3D coordinates with text lines, UV maps with document layouts, or foreground masks with text lines. In fact, 3D coordinates represent the position of pixels in three-dimensional space, while UV maps accurately map each point in the image to the surface of a 3D object on a 2D plane. These two global features also contain document foreground information similar to foreground masks. However, existing methods ignore the feature complementarity between global granularities. 
Second, fine-grained information (text lines and layouts) is adopted to improve the readability of distorted document images. Yet most existing methods lack consideration of the mutual constraints between the horizontal and vertical directions of the document.
Third, while addressing both global and local granularities, most previous works have not explored the feature correlation between multiple tasks.

We propose an end-to-end multi-granular feature adaptive constraint and geometry-aware network named SalmRec, focusing on document deformation morphology and text recognition. SalmRec is inspired by multi-task feature fusion and introduces three-dimensional (3D) and plane coordinates (UV maps) globally. The local information is inspired by the order in which humans read documents. Horizontal and vertical line features representing the X and Y directions are used to constrain fine-grained features. Furthermore, adaptive geometric perception is performed across multiple tasks of distorted image features to enhance feature interactions and mutual constraints. A gating mechanism is adopted to enable effective feature selection on global and local features respectively. The contributions of our work are summarized as follows:

\begin{itemize}
\item{We propose an end-to-end multi-task learning network based on inter-granular feature fusion for distorted document rectification, incorporating global and local, 2D and 3D coordinate features to improve the dewarping effect and text recognition accuracy.}
\item{We introduce an inter-task feature aggregation module to promote task complementarity between multi-granularity, and a gating mechanism to dynamically weight global and local features to enhance geometric representation learning of distorted images.}
\item{Our method achieves state-of-the-art performance on three public benchmarks. Extensive experiments verify the role of different granular features on the dewarping performance. The ablation study demonstrates the effectiveness of the feature aggregation module and the gating mechanism.}
\end{itemize}

\section{Related Work}

In the past few years, many excellent methods have promoted the performance improvement of the correction task. Given that the potential features of multiple distorted images (3D coordinates, UV map, horizontal and vertical lines) are mined during the synthesis of training data. These works have also gradually transitioned from single-task to multi-task learning. The introduction of auxiliary tasks has been proven to be more effective in this task.

\subsection{Traditional Methods}
Before the advent of deep learning, traditional document image rectification methods usually adopt parametric regression and rely on easily observable features on the image surface, such as text lines \cite{Huang2015TextLE}, cylindrical surface curves \cite{Wada1997ShapeFS, Courteille2007ShapeFS, Liang2008GeometricRO, Meng2018ExploitingVF} and extracting clear document boundaries \cite{Brown2006GeometricAS, Tsoi2007MultiViewDR}. These methods focus on shallow visual representations in document images and do not need to reconstruct 3D shapes. When encountering dark lighting, shadow coverage, and close background and foreground colors, these rule-based processing methods are powerless in real scenes.

\subsection{Deep Learning Based Methods}
{\bf{Without Auxiliary Information Methods.}} Methods without other auxiliary information guidance \cite{Ma2018DocUNetDI, Liu2020GeometricRO, Xie2022DocumentDW, Xue2022FourierDR, Yu_2024_WACV} directly predicted the two-dimensional deformation field. DocUNet \cite{Ma2018DocUNetDI} utilized two stacked UNet structures for training on synthetic datasets. AGUN \cite{Liu2020GeometricRO} designed a pyramid encoder-decoder network to predict anti-distorted meshes at multiple resolutions in a coarse-to-fine manner. DDCP \cite{Xie2022DocumentDW} rectified distorted document images by estimating the mapping between sparse control points and reference points. FDRNet \cite{Xue2022FourierDR} focused on high-frequency components in Fourier space to capture structural information in documents. DocReal \cite{Yu_2024_WACV} proposed the Attention Enhanced Control Point (AECP) module to better capture local deformations and releases a Chinese benchmark for follow-up research.

{\bf{With Auxiliary Information Methods.}}
Recent work focuses on introducing additional information related to document features to assist correction \cite{Das2019DewarpNetSD, Feng2021DocTrDI, 10374269, Liu2023DocMAEDI, Feng2022GeometricRL, Jiang2022RevisitingDI, Dai2023MataDocMA, li2023foreground, Feng2021DocScannerRD, Zhang2022MariorMR, li2023layout, UVDoc, Tang2024EfficientJR, Kumari2024AmIR}.
In terms of focusing on \textit{global features}, some methods \cite{Feng2021DocTrDI,Feng2022GeometricRL, Zhang2022MariorMR, Feng2021DocScannerRD, Tang2024EfficientJR, Kumari2024AmIR} remove background as a preprocessing stage. Often a lightweight network was used to predict the binary mask and then multiply it with the input image to obtain an image with only the document foreground as the input of the rectification network.
Several previous works \cite{Das2019DewarpNetSD, Feng2022GeometricRL} introduce 3D coordinate features as auxiliary tasks. DewarpNet \cite{Das2019DewarpNetSD} released a large-scale training set Doc3D and proposed a two-stage network to predict 3D coordinates and backward maps respectively, where 3D coordinates are used as the input of the second-stage network. DocGeoNet \cite{Feng2022GeometricRL} directly concatenated 3D coordinate features with text line features as the input of the decoder to predict the 2D deformation field coordinates. LA-DocFlatten \cite{li2023layout} proposed to jointly regress UV texture coordinates and document layout information for auxiliary correction.

In terms of focusing on \textit{local features}, unlike the method of removing the document background, RDGR \cite{Jiang2022RevisitingDI} only predicted the 2D deformation field coordinates of the document foreground boundary, and constrained the 2D grid regularization through the pre-trained text line detection network. FTDR \cite{li2023foreground} used the cross attention mechanism to make the model focus on the foreground and text line areas at the same time. Our proposed SalmRec promotes the model to characterize the commonalities and differences between different granularities through collaborative learning among multiple tasks.

\begin{figure*}[!t]
\centering
\includegraphics[width=0.9\linewidth]{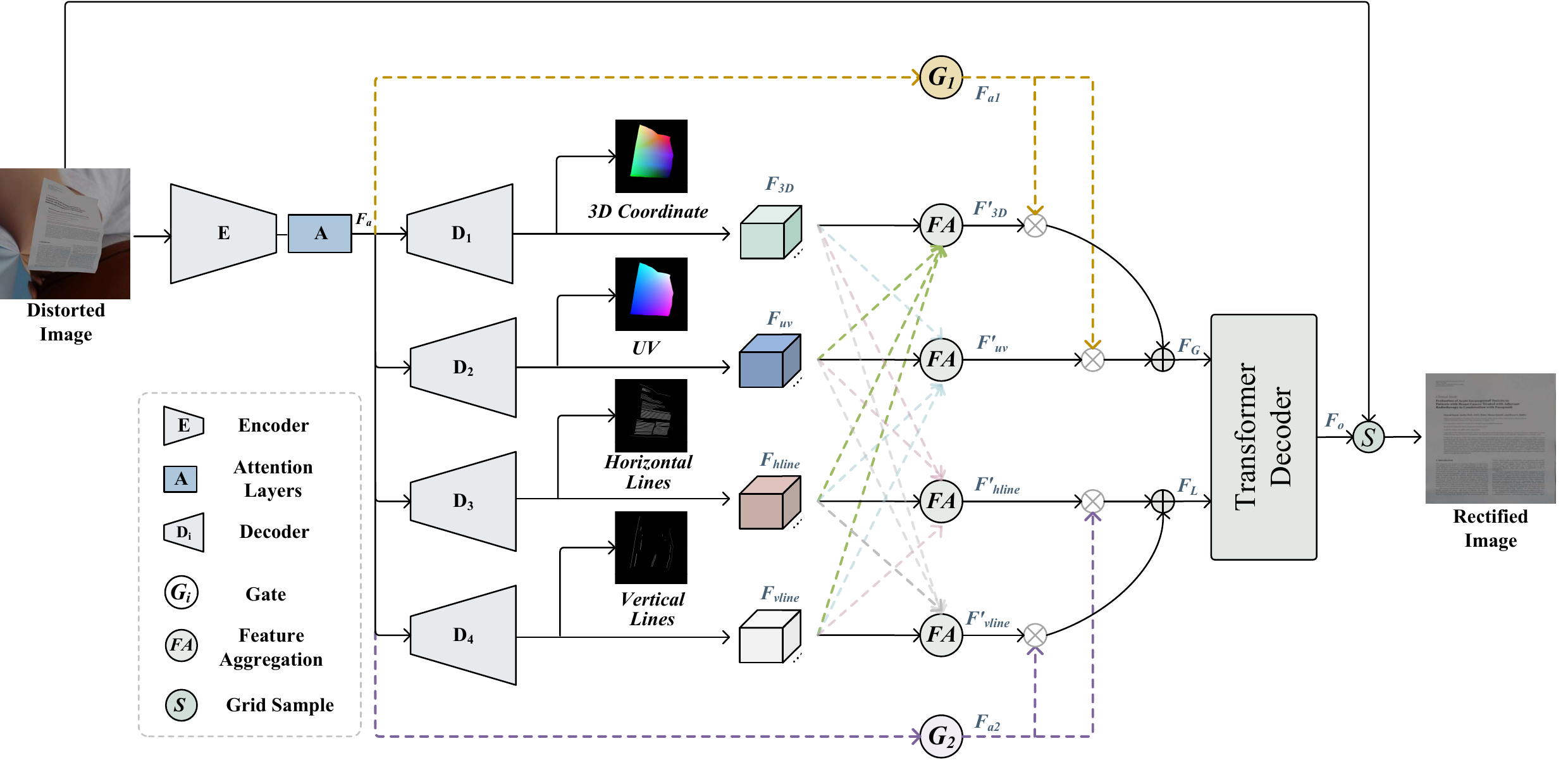}
\caption{{\bf Overview architecture of our proposed SalmRec.} For a given \textit{distorted image}, SalmRec learns to predict a 2D deformation field by multi-decoder segmentation module ($D_i$), inter-task feature aggregation ($FA$), gating mechanisms ($G_i$), and Transformer decoder. $\boldsymbol{F}$ represents the output feature map.}
\label{model_overview}
\end{figure*}

\section{Methodology}
The proposed SalmRec mainly consists of several parts, including \textit{Multi-decoder Segmentation Module}, the \textit{Inter-task Feature Aggregation Module (FA)}, the \textit{Gating Module} and the \textit{Transformer Decoder}. The overall structure of our model is shown in Fig. \ref{model_overview}.

{\bf Network Architecture.} The multi-decoder segmentation module shares the encoder and segments the features of four different granularities under the guidance of multiple tasks($E$, $A$ and $D_i$ in Fig. \ref{model_overview}). 
The inter-task feature aggregation module adopts leave-one-out combination to learn the correlation between tasks (\textit{FA} in Fig. \ref{model_overview}). 
The gating module ($G_i$ in Fig. \ref{model_overview}) selects the importance of global and local information respectively, highlighted with yellow and purple dotted lines. 
Then the outputs of the two gating modules are concatenate in the channel dimension and fed into the multi-layer Transformer decoder to predict the 2D deformation field coordinates. Finally, the coordinates are used to de-distort the input warped image.

{\bf Multi-decoder Segmentation}. The entire segmentation module adopts the UNet \cite{ronneberger2015u} structure. 
For a given distorted document image $\boldsymbol{I} \in \mathbb {R} ^ {H_{in}\times W_{in}\times C_{in}}$, it is first resized to $\boldsymbol{I_r} \in \mathbb {R} ^ {H_r\times W_r\times C_r}$, where $H_r$=$W_r$=$448$, and $C_r$=$C_{in}$=$3$ represents the number of RGB channels of the original image. 
Then $\boldsymbol{I_r}$ is fed into the visual encoder $E$ as the input of our rectification network. 
The encoder contains five layers in total. 
Two $3\times3$ Convolutions (Conv), Batch Normalization (BN) \cite{Ioffe2015BatchNA} and Rectified Linear Unit (ReLU) activation \cite{Nair2010RectifiedLU} blocks (ConvBNReLU, CBR) are used to extract the general representation of the input image $\boldsymbol{I_r}$ in first layer. 
The remaining four layers are downsampling modules. It consists of a maximum pooling and two ConvBNReLU.
$L_a$ layers of self-attention are introduced before the decoder to learn the long-distance dependencies between feature map elements, where $L_a$=$6$. The feature $\boldsymbol{F}_a$ is obtained as the input of decoders.

Multiple decoders are learned independently. 
Each decoder has four layers with exactly the same structure, including a bilinear interpolation upsampling and two ConvBNReLU. The channels of the four feature maps in the decoder are \{128, 96, 48, 32\}.
For each layer in the decoder, a task regression module predicts the feature map of a specific task.
For each decoder, the feature maps of the four layers are resized to one-sixteenth of the input image $\boldsymbol{I_r}$, concatenated in the channel dimension, and input into a layer of ConvBNReLU for channel reduction. 
The feature maps of the four tasks are obtained respectively ($\boldsymbol{F}_{3D}$, $\boldsymbol{F}_{uv}$, $\boldsymbol{F}_{hline}$, $\boldsymbol{F}_{vline} \in \mathbb {R} ^ {H\times W\times C}$  in Fig. \ref{model_overview}, where $H$=$W$=${H_r}/{16}$, and $C$=$256$).
For better visualization, we omit the skip connection lines between the encoder and decoder in Fig. \ref{model_overview}.

\begin{figure}[!t]
\centering
\includegraphics[]{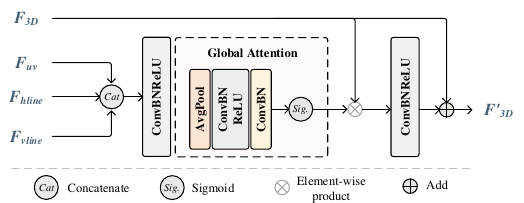}
\caption{Inter-task Feature Aggregation (FA) Module. The four tasks are divided into four groups, where $\boldsymbol{F}_{3D}$ is fused with the global attention score obtained from the other three tasks for feature interconnection aggregation.}
\label{FA}
\end{figure}

{\bf Inter-task Feature Aggregation Module (\textit{FA}).} We introduce the global attention mechanism \cite{Liu2021GlobalAM}, divide the feature maps of the four tasks into four groups for feature aggregation using a leave-one-out combination, learn the relationship between tasks, and reduce the interference of redundant features. The module structure is illustrated in Fig. \ref{FA}. Here we take the feature of purifying 3D coordinates ($\boldsymbol{F}_{3D}$) as an example. First, concatenate the other three feature maps ($\boldsymbol{F}_{uv}$, $\boldsymbol{F}_{hline}$, $\boldsymbol{F}_{vline}$). Then utilize a layer of ConvBNReLU (CBR) to fuse them and reduce the number of channels. This map implies information about different perspectives of the distorted image. The global representation of this mixed feature map is obtained by integrating pooling, convolution, and Sigmoid operations, and multiplied by $\boldsymbol{F}_{3D}$ to obtain the refined global interaction representation. The calculation formulas of FA module are as follows:
\begin{equation}
\begin{aligned}
  &\boldsymbol{F}_{cat}=CBR(CAT(\boldsymbol{F}_{uv}, \boldsymbol{F}_{hline}, \boldsymbol{F}_{vline})) \\
  &\boldsymbol{F}_{{3D}}^{\prime}=CBR(GlobalAttention(\boldsymbol{F}_{cat}) \times \boldsymbol{F}_{3D}) + \boldsymbol{F}_{3D} \\
\end{aligned}
\end{equation}
\noindent where $CBR(\cdot)$ is ConvBNReLU block, and $CAT(\cdot)$ is the concatenation operation in the channel dimension.

\begin{figure}[!t]
\centering
\includegraphics[]{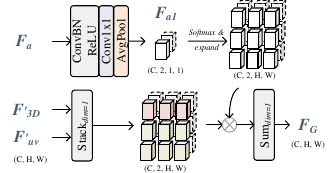}
\caption{Gating Mechanism. $\boldsymbol{F}_a$ is the universal feature extracted from the original distorted image by the encoder of the segmentation module. $\boldsymbol{F}^{\prime}_{3D}$ and $\boldsymbol{F}^{\prime}_{uv}$ are two aggregated features with global properties. Similar operations are performed on local features $\boldsymbol{F}^{\prime}_{hline}$ and $\boldsymbol{F}^{\prime}_{vline}$. $C$, $H$, $W$ represents the channel, height and width of the feature map respectively.}
\label{Gate}
\end{figure}

{\bf Gating Module.} Inspired by routing-based multi-task learning \cite{Strezoski2019ManyTL, Conde2024InstructIRHI}, we design a gating mechanism to refine global and local tasks respectively, and further adaptively guide the selection of features of different granularities. 
The schematic diagram of this module is demonstrated in Fig. \ref{Gate}. 
Here we take learning to adjust the weights of global features as an example ($G_1$ in Fig. \ref{model_overview}). 
For the global features $\boldsymbol{F}^{\prime}_{3D}$ and $\boldsymbol{F}^{\prime}_{uv}$ that represent the foreground of the document output by the aggregation module, two feature maps are spliced in the new dimension. 
The output $\boldsymbol{F}_a$ of the segmentation module is the original feature representation of the distorted image. The feature weights corresponding to $\boldsymbol{F}^{\prime}_{3D}$ and $\boldsymbol{F}^{\prime}_{uv}$ are obtained through Convolution, Pooling and Softmax operations. 
Similar calculations are also performed for the feature maps $\boldsymbol{F}^{\prime}_{hline}$ and $\boldsymbol{F}^{\prime}_{vline}$ representing local information.
\begin{equation}
\begin{aligned}
  &\boldsymbol{F}_{a1}=Squeeze(\boldsymbol{F}_a) \\
  &\boldsymbol{F}_{{a1}}^{\prime}=Expand(Softmax(\boldsymbol{F}_{a1})) \\
  &\boldsymbol{F}_{G}=Stack(\boldsymbol{F}_{{3D}}^{\prime}, \boldsymbol{F}_{{uv}}^{\prime})_{dim=1} \\
  &\boldsymbol{F}_{G}=\boldsymbol{F}_{{a1}}^{\prime} \times \boldsymbol{F}_{G}
\end{aligned}
\end{equation}
\noindent where $Squeeze$ is an operation that compresses the feature map dimension, including $CBR$, 1$\times$1 convolution, and AvgPool.

{\bf Transformer Decoder}. After the global and local elements are purified by the gating mechanism, the two features are directly concatenated and fed into the Transformer decoder. The decoder consists of a multi-layer multi-head self-attention mechanism and feed-forward. The grid sampling module proposed by DocTr \cite{Feng2021DocTrDI} is used to upsample and output two-dimensional grid coordinates ($\hat{B}$). 

{\bf Loss Function.} For the global feature 3D coordinates and UV map, we use Mean Square Error (MSE) loss to supervise the training process ($\mathcal{L}_{\textit{3D}}$ and $\mathcal{L}_{\textit{UV}}$). 
Predict local information horizontal and vertical lines accompanied by Binary Cross Entropy (BCE) \cite{Boer2005ATO} loss. The formulas are as follows:
\begin{equation}
\begin{aligned}
  &\mathcal{L}_{\textit{bce}}^{h}=-\frac{1}{K}\sum_{k=1}^{\textit{K}}[y^{h}_{k} \log(\hat{y}^{h}_{k})+(1 - y^{h}_{k}) \log(1 - \hat{y}^{h}_{k})] \\
  &\mathcal{L}_{\textit{bce}}^{v}=-\frac{1}{K}\sum_{k=1}^{\textit{K}}[y^{v}_{k} \log(\hat{y}^{v}_{k})+(1 - y^{v}_{k}) \log(1 - \hat{y}^{v}_{k})] \\
\end{aligned}
\end{equation}
\noindent
where $K$ is the pixel count of the input image, $h$ and $v$ represent horizontal and vertical lines, respectively. $y$ and $\hat{y}$ are the ground truth and predicted results.

In order to balance the difference in the number of pixels between the background and the foreground in the local features, we introduce the line loss ($\mathcal{L}_{\textit{line}}^{h}$ and $\mathcal{L}_{\textit{line}}^{v}$) proposed by RDGR \cite{Jiang2022RevisitingDI}. 
This loss uses the Mean Square Error weighted pixel ratio. ‌In the cause of weighing the output of different layers of the decoder in the segmentation module, each layer is assigned a different loss weight. The losses are calculated as follows:
\begin{equation}
\begin{aligned}
  &\mathcal{L}_{\textit{pos}}^{h}=\frac{1}{K_{pos}}\sum_{pos=1}^{\textit{$K_{pos}$}}{(y^{h}_{pos}-\hat{y}^{h}_{pos}})^{2} \\
  &\mathcal{L}_{\textit{neg}}^{h}=\frac{1}{K_{neg}}[(\sum_{k=1}^{\textit{K}}{(y^{h}_{k}-\hat{y}^{h}_{k}})^{2}) - (\sum_{pos=1}^{\textit{$K_{pos}$}}{(y^{h}_{pos}-\hat{y}^{h}_{pos}})^{2})] \\
  &\mathcal{L}_{\textit{line}}^{h}=\frac{1}{K}({K_{pos}}\mathcal{L}_{\textit{neg}}^{h} + {K_{neg}}\mathcal{L}_{\textit{pos}}^{h})
\end{aligned}
\end{equation}
\noindent
Where $pos$ means Positive Pixels and $K_{pos}$ represents the number of pixels in the mask part of the horizontal line. The remaining pixels are Negative and are recorded as $neg$. The loss $\mathcal{L}_{\textit{line}}^{v}$ of vertical lines is calculated in the same way. 
The overall loss of the segmentation module ($\mathcal{L}_{\textit{seg}}$) is calculated as follows:
\begin{equation}
\begin{aligned}
  &\mathcal{L}_{\textit{line}}=\sum_{i=1}^{\textit{L}}[\frac{1}{2{L}-i}(\mathcal{L}_{\textit{bce}}^{h} + \mathcal{L}_{\textit{bce}}^{v}) + \mathcal{L}_{\textit{line}}^{h} + \mathcal{L}_{\textit{line}}^{v}] \\
  &\mathcal{L}_{\textit{seg}}=\mathcal{L}_{\textit{3D}} + \mathcal{L}_{\textit{UV}} + \mathcal{L}_{\textit{line}}
\end{aligned}
\end{equation}
\noindent
where $L$ is the number of layers of the decoder. $i$ is the $i$-$th$ layer of each decoder.

$L_1$ loss is utilized to calculate the distance between the grid coordinates ($\hat{B}$) and the label ($B$), the loss formula is:
\begin{equation}
    \mathcal{L}_\textit{b} = \left \| B - \hat{B} \right \| _\bold{1}
\end{equation}

The overall rectification loss is as follows:
\begin{equation}
    \mathcal{L} = \lambda \mathcal{L}_\textit{b} + \mathcal{L}_{\textit{seg}}
\end{equation}
where $\lambda$ represents a hyper-parameter weight, which is assigned a value of 5 to ensure an balance among various loss functions.

\section{EXPERIMENTS}
\subsection{Datasets}
The public distorted image training set used by most methods is Doc3D \cite{Das2019DewarpNetSD}. 
However, this dataset lacks fine-grained annotations (text lines, horizontal or vertical lines). Previous methods RDGR \cite{Jiang2022RevisitingDI}, DocGeoNet \cite{Feng2022GeometricRL} and FTDR \cite{li2023foreground} use threshold binarization or pre-trained line detection models to obtain fine-grained results, which inevitably introduce inaccuracies. 
In our paper, the distorted document image dataset DocDewarpHV \footnote{https://github.com/xiaomore/DocDewarpHV} with accurate annotations is used, which also has 110,000 images. 
In addition to 3D coordinates and UV maps, each distorted image is annotated with two types of local distortion information: horizontal and vertical lines. We use the following three benchmarks as test sets.

{\bf DIR300 Benchmark.} This evaluation dataset is proposed by the paper DocGeoNet \cite{Feng2022GeometricRL}. This is the largest and most diverse benchmark. It contains 300 distorted images taken in natural scenes with different intensities of lighting, shadows, folds, curves, and various background interferences. 

{\bf DocReal Benchmark.} Compared with the DIR300 and DocUNet benchmarks, DocReal \cite{Yu_2024_WACV} is the largest Chinese evaluation dataset, containing 200 document images, covering a variety of types such as contracts, bills and textbooks.

{\bf DocUNet Benchmark.} The benchmark proposed by DocUNet \cite{Ma2018DocUNetDI} is the first application of deep learning on this task. This benchmark has been widely used in recent studies and contains 130 distorted images of documents in natural environments captured using mobile devices. The dataset covers a variety of document types, including bills, flyers, books and posters.

\subsection{Evaluation Metrics}
{\bf MS-SSIM, LD and AD.} Our evaluation uses metrics widely adopted by previous related works \cite{Das2019DewarpNetSD, Feng2021DocTrDI, 10374269, Liu2023DocMAEDI, Feng2022GeometricRL, Jiang2022RevisitingDI, li2023foreground, Feng2021DocScannerRD, Zhang2022MariorMR, li2023layout, UVDoc, Tang2024EfficientJR, Kumari2024AmIR}. Multi-Scale Structural Similarity (MS-SSIM) \cite{Wang2004ImageQA} is used to evaluate the global similarity between the rectified image and the reference image. Another metric, Local Distortion (LD) \cite{You2016MultiviewRO}, focuses on local details by calculating SIFT flow \cite{Liu2011SIFTFD}. Aligned Distortion (AD) \cite{Ma2022LearningFD} optimizes the limitations of MS-SSIM, which is insensitive to subtle global changes, and LD, which tends to inaccurately mark errors in regions lacking texture.

{\bf ED and CER.} For the evaluation of text recognition accuracy on rectified images, we follow previous methods and use Edit Distance (ED) and Character Error Rate (CER). ED defines the minimum number of times a string is transformed into a reference string after insertion, deletion, and substitution operations. The Character Error Rate (CER) is calculated by dividing the result of ED by the length of the reference string. To align with existing methods, 90 rich text document images are evaluated on the DIR300 benchmark. On the DocUNet benchmark, the text recognition accuracy is evaluated on 50 and 60 images, respectively. On the Chinese benchmark DocReal, the ED and CER results of all 200 images are displayed.

\subsection{Implementation Details}
Our entire network is implemented in the Pytorch \cite{Paszke2017AutomaticDI} deep learning framework. It is trained on four 32GB NVIDIA V100 GPUs for 60 epochs with batch size 32 and 10,000 warmup steps. The optimizer is AdamW \cite{Loshchilov2017DecoupledWD}. The learning rate adopts the cosine decay strategy with a maximum learning rate of $1.2\times10^{-4}$ and a minimum of $5\times10^{-7}$. Following previous works \cite{Feng2021DocTrDI, 10374269, Liu2023DocMAEDI, Feng2022GeometricRL, Jiang2022RevisitingDI, li2023foreground, Feng2021DocScannerRD, Zhang2022MariorMR, li2023layout}, the evaluation metrics MS-SSIM, LD, and AD are evaluated by Matlab R2019a. The OCR performance is evaluated on two English benchmarks using Tesseract v5.0.1.2022011 \cite{Smith2007AnOO} with pytesseract v0.3.8 \footnote{https://pypi.org/project/pytesseract/}. Given PaddleOCR's \footnote{https://github.com/PaddlePaddle/PaddleOCR} excellent performance in Chinese text recognition, it is used to measure CER and ED on the Chinese DocReal benchmark, where the text detection model is DBNet \cite{9726868} with version ``detv4\_teacher\_inference'' and text recognition model is SVTR\_LCNet \cite{Du2022SVTRST} with version ``OCRv4\_rec\_server\_infer''.

\begin{table}[t]\rmfamily
\centering
  \caption{Comparisons rectification performance on DIR300 benchmark \cite{Feng2022GeometricRL}. ``$\uparrow$'' indicates the higher the better and ``$\downarrow$'' denotes the opposite. The best performing result is shown in \textbf{bold} font, and the second best result is shown with an \underline{underline}.}
  \label{tab:eval_dir300}
  \begin{tabular}{cccc}
    \toprule
    Method & MS-SSIM$\uparrow$ & LD$\downarrow$/AD$\downarrow$ & ED$\downarrow$/CER$\downarrow$ \\
    \midrule
    Distorted & 0.32 & 39.58/0.771 & 1500.56/0.5234 \\
    DocProj \cite{Li2019DocumentRA} & 0.32 & 30.63/- & 958.89/0.3540 \\
    DewarpNet \cite{Das2019DewarpNetSD} & 0.49 & 13.94/0.331 & 1059.57/0.3557 \\
    DocTr \cite{Feng2021DocTrDI} & 0.62 & 7.21/0.254 & 699.63/0.2237 \\
    DDCP \cite{Xie2022DocumentDW} & 0.55 & 10.95/0.357 & 2084.97/0.5410 \\
    DocGeoNet \cite{Feng2022GeometricRL} & 0.64 & 6.40/0.242 & 664.96/0.2189 \\
    PaperEdge \cite{Ma2022LearningFD} & 0.58 & 8.00/0.255 & 508.73/0.2069 \\
    FTDR \cite{li2023foreground} & 0.61 & 7.68/0.244 & 652.80/0.2115 \\
    DocScanner \cite{Feng2021DocScannerRD} & 0.62 & 7.06/0.225 & 562.72/0.1943 \\
    LA-DocFlatten \cite{li2023layout} & 0.65 & \underline{5.70}/\underline{0.195} & 511.13/0.1891 \\
    UVDoc \cite{UVDoc} & 0.62 & 7.71/0.218 & 605.24/0.2601 \\
    DocRes \cite{Zhang2024DocResAG} & 0.63 & 6.81/0.243 & 764.20/0.2403 \\
    DocTLNet \cite{Kumari2024AmIR} & \underline{0.66} & 5.75/- & \underline{482.57}/\underline{0.1767} \\
    Ours & \textbf{0.67} & \textbf{5.14}/\textbf{0.178} & \textbf{444.07}/\textbf{0.1400}
 \\
    \bottomrule
  \end{tabular}
\end{table}

\begin{figure*}[!t]
\centering  
\subfloat{\includegraphics[scale=0.3]{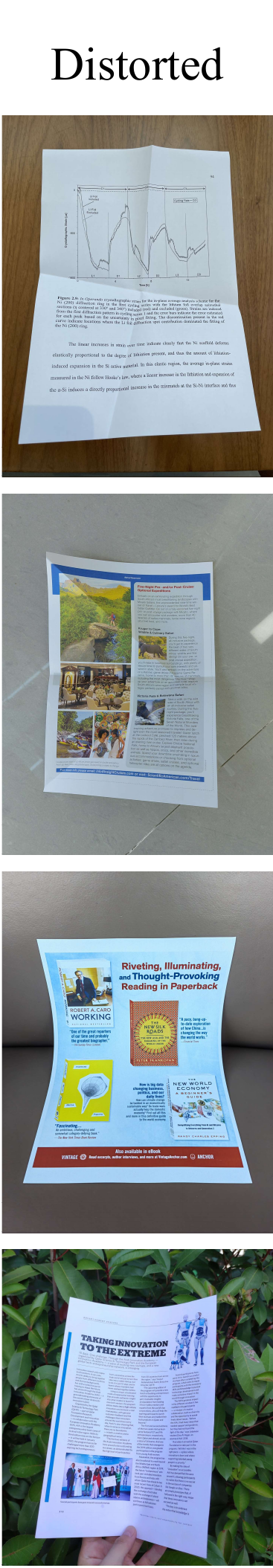}}
\subfloat{\includegraphics[scale=0.3]{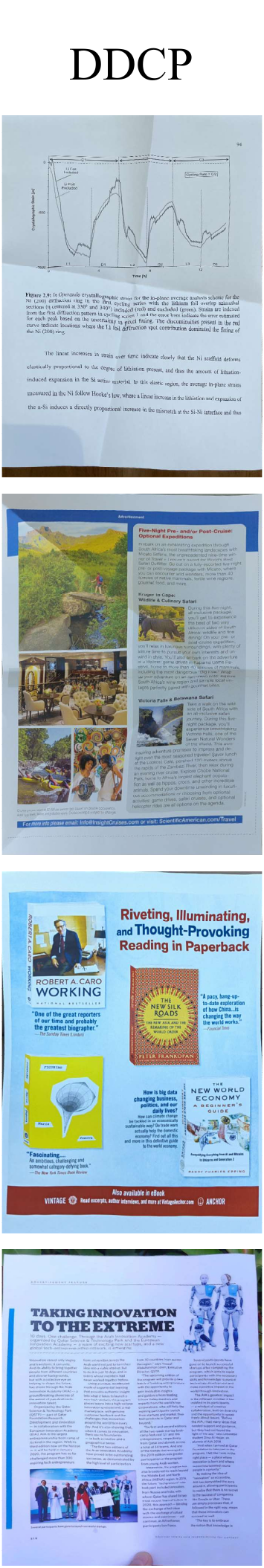}}
\subfloat{\includegraphics[scale=0.3]{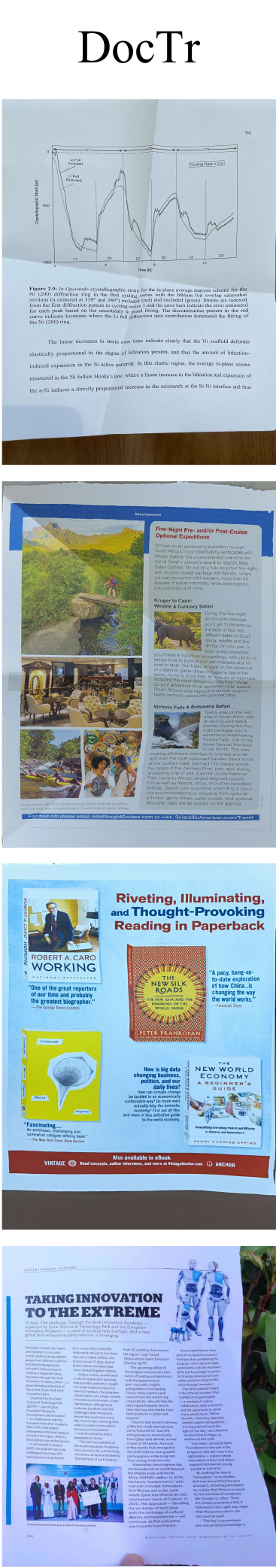}}
\subfloat{\includegraphics[scale=0.3]{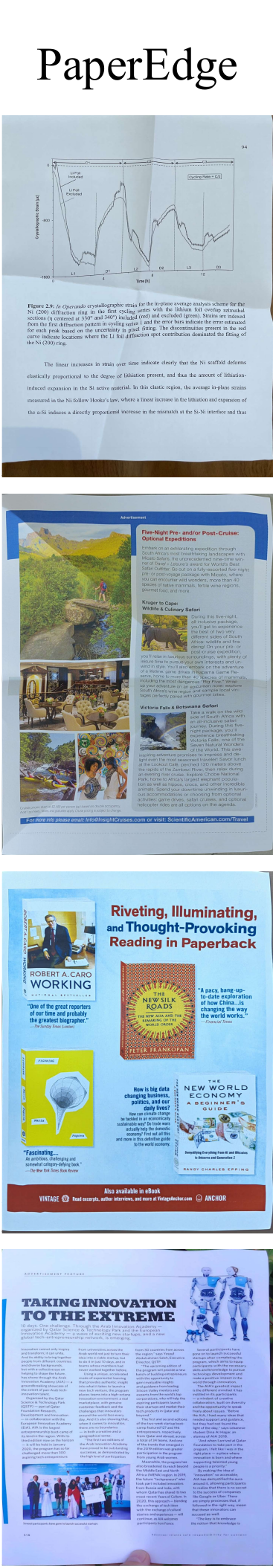}}
\subfloat{\includegraphics[scale=0.3]{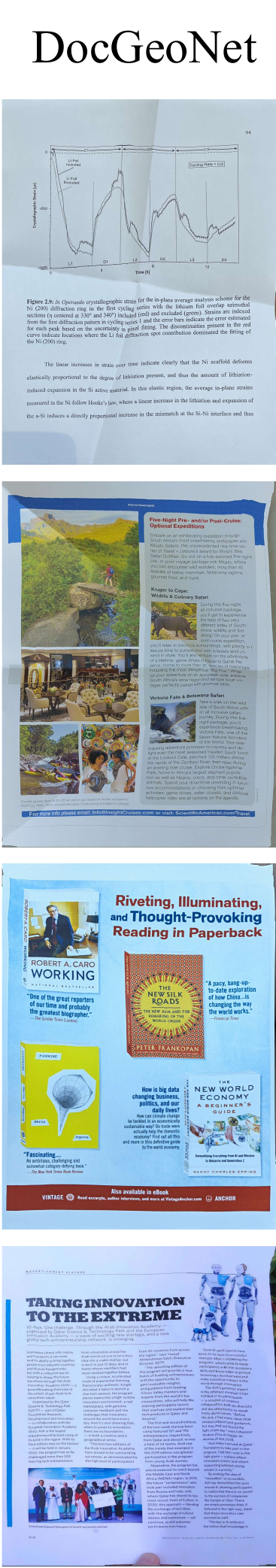}}
\subfloat{\includegraphics[scale=0.3]{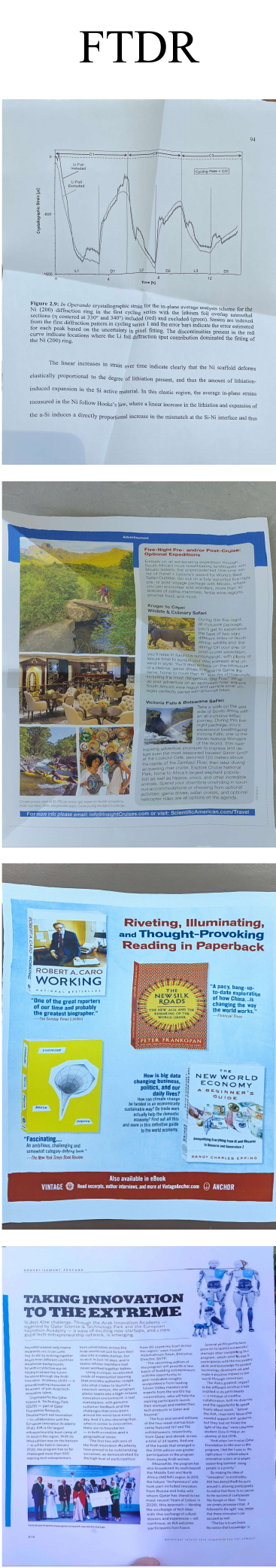}}
\subfloat{\includegraphics[scale=0.3]{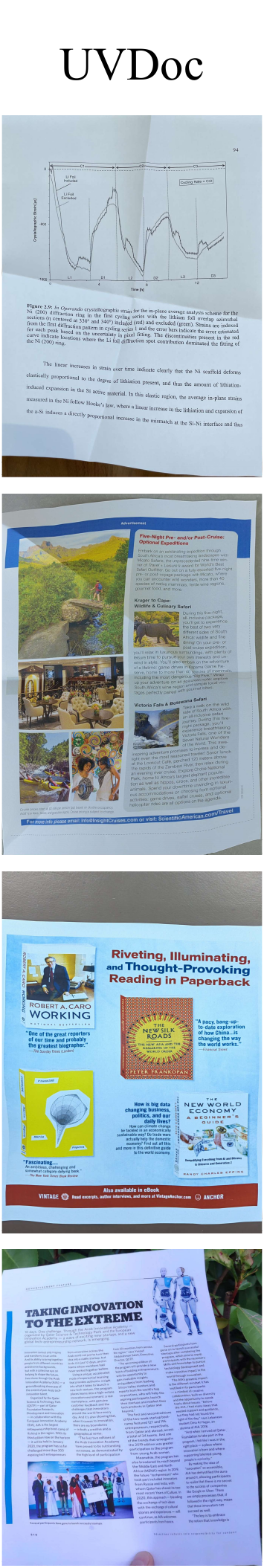}}
\subfloat{\includegraphics[scale=0.3]{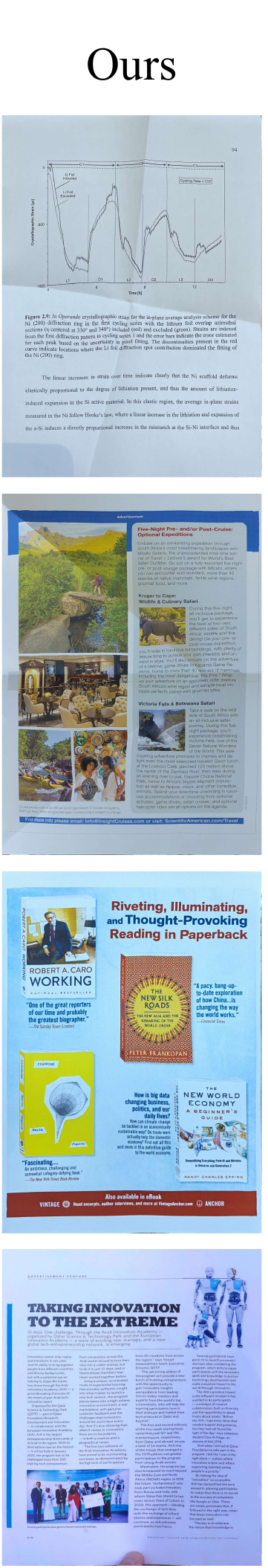}}
\subfloat{\includegraphics[scale=0.3]{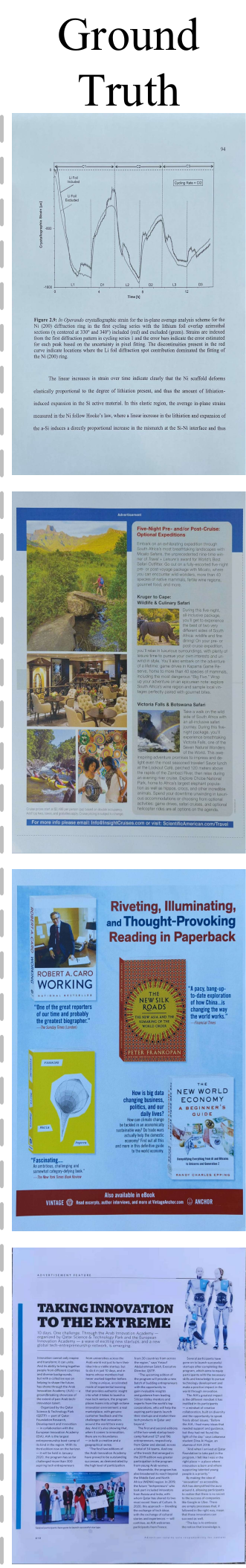}}
\caption{Qualitative visual comparison with existing methods on DIR300 benchmark \cite{Feng2022GeometricRL}. The first column ``Distorted'' means the given distorted document image, and the last column ``Ground-Truth'' is the flattened reference image.}
\label{DIR300_vis}
\end{figure*}

\begin{table}[t]\rmfamily
\centering
  \caption{Comparisons rectification performance on DocReal benchmark \cite{Yu_2024_WACV}. The data in this table evaluated by ourself from the public code or rectified results of previous methods.}
  \label{tab:eval_DocReal}
  \begin{tabular}{cccc}
    \toprule
    Method & MS-SSIM$\uparrow$ & LD$\downarrow$/AD$\downarrow$ & ED$\downarrow$/CER$\downarrow$ \\
    \midrule
    Distorted & 0.32 & 35.79/0.771 & 1500.56/0.5234 \\
    DocTr & 0.55 & 12.68/0.322 & 208.51/0.2943 \\
    DDCP & 0.47 & 16.36/0.449 & 264.71/0.3300 \\
    DocGeoNet & 0.55 & 12.24/0.314 & 216.68/0.2974 \\
    PaperEdge & 0.52 & 11.47/0.303 & \underline{186.86}/\textbf{0.2451} \\
    RDGR & 0.54 & 11.47/0.333 & 188.00/0.2768 \\
    FTDR & 0.51 & 13.25/0.350 & 217.68/0.2890 \\
    DocScanner & 0.54 & 12.36/0.308 & 196.14/0.2709 \\
    UVDoc & 0.55 & 11.40/0.274 & 203.12/0.2688 \\
    DocRes & 0.55 & 11.71/0.330 & 224.44/0.3087 \\
    DocReal & \underline{0.56} & \underline{9.83}/\underline{0.238} & \textbf{184.54}/\underline{0.2485} \\
    Ours & \textbf{0.59} & \textbf{8.41}/\textbf{0.229} & 193.06/0.2604 \\
    \bottomrule
  \end{tabular}
\end{table}

\begin{table*}[\tblwidth]\rmfamily
\centering
  \caption{Comparisons on DocUNet benchmark \cite{Ma2018DocUNetDI}. ``$\uparrow$'' indicates the higher the better and ``$\downarrow$'' denotes the opposite. The best performing result is shown in \textbf{Bold} font, and the second best result is shown with an \underline{underline}.}
  \label{tab:eval_docunet}
  \begin{tabular}{llccccc}
    \toprule
    Method & Venue & MS-SSIM $\uparrow$ & LD $\downarrow$ & AD $\downarrow$ & ED $\downarrow$ & CER $\downarrow$\\
    \midrule
    Distorted & - & 0.25 & 20.51 & 1.012 & 2111.56/1552.22 & 0.5352/0.5089 \\
    DocUNet \cite{Ma2018DocUNetDI} & CVPR’18 & 0.41 & 14.19 & - & 1933.66/1259.83 & 0.4632/0.3966 \\
    DocProj \cite{Li2019DocumentRA} & TOG’19 & 0.29 & 18.01 & 0.994 & 1712.48/1165.93 & 0.4267/0.3818 \\
    DewarpNet \cite{Das2019DewarpNetSD} & ICCV’19 & 0.47 & 8.39 & 0.426 & 885.90/525.45 & 0.2373/0.2102 \\
    FCN-based \cite{Xie2020DewarpingDI} & DAS’20 & 0.45 & 7.84 & 0.434 & 1792.60/1031.40 & 0.4213/0.3156 \\
    Piece-Wise \cite{Das2021EndtoendPU} & ICCV’21 & 0.49 & 8.64 & 0.468 & 1069.28/743.32 & 0.2677/0.2623 \\
    DocTr \cite{Feng2021DocTrDI} & ACM MM’21 & 0.51 & 7.76 & 0.396 & 724.84/464.83 & 0.1832/0.1746 \\
    DDCP \cite{Xie2022DocumentDW} & ICDAR’21 & 0.47 & 8.99 & 0.453 & 1442.84/745.35 & 0.3633/0.2626 \\
    FDRNet \cite{Xue2022FourierDR} & CVPR’22 & \underline{0.54} & 8.21 & - & 829.78/514.90 & 0.2068/0.1846 \\
    RDGR \cite{Jiang2022RevisitingDI} & CVPR’22 & 0.50 & 8.51 & 0.461 & 729.52/420.25 & 0.1717/0.1559 \\
    DocGeoNet \cite{Feng2022GeometricRL} & ECCV’22 & 0.50 & 7.71 & 0.380 & 713.94/379.00 & 0.1821/0.1509 \\
    PaperEdge \cite{Ma2022LearningFD} & SIGGRAPH’22 & 0.47 & 7.99 & 0.392 & 777.76/375.60 & 0.2014/0.1541 \\
    FTDR \cite{li2023foreground} & ICCV'23 & 0.50 & 8.43 & 0.376 & 697.52/450.92 & 0.1705/0.1679 \\
    DocScanner \cite{Feng2021DocScannerRD} & arXiv'23 & 0.52 & 7.45 & 0.334 & \textbf{632.30}/390.40 & \underline{0.1650}/0.1490 \\
    DocTr-Plus \cite{10374269} & TMM'23 & 0.51 & 7.52 & - & -/447.47 & -/0.1695 \\
    LA-DocFlatten \cite{li2023layout} & TOG'23 & 0.53 & \underline{6.72} & \underline{0.300} & 695.00/391.90 & 0.1750/0.1530 \\
    UVDoc \cite{UVDoc} & SIGGRAPH'23 & \textbf{0.55} & 6.79 & 0.310 & 797.92/493.13 & 0.1975/0.1611 \\
    DocRes \cite{Zhang2024DocResAG}	& CVPR'24 & 0.47 & 9.37 & 0.471 & 912.50/500.27 & 0.2406/0.1751 \\
    DocTLNet \cite{Kumari2024AmIR} & IJDAR'24 & 0.51 & \textbf{6.70} & - & -/377.12 & -/0.1504 \\
    DocReal \cite{Yu_2024_WACV} & WACV'24 &0.50 & 7.03 & \textbf{0.286} & 730.96/\underline{360.32} & 	0.1909/\underline{0.1443} \\
    Ours & - & 0.51	& 7.10 & 0.310 & \underline{638.34}/\textbf{330.15} & \textbf{0.1619}/\textbf{0.1326} \\
    \bottomrule
  \end{tabular}
\end{table*}

\begin{figure}[t]
\centering
\subfloat{\includegraphics[scale=0.3]{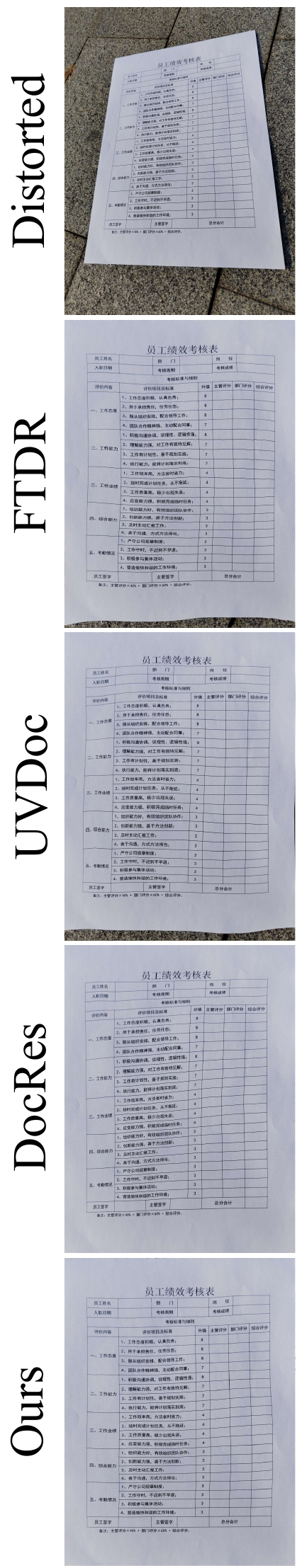}}
\subfloat{\includegraphics[scale=0.3]{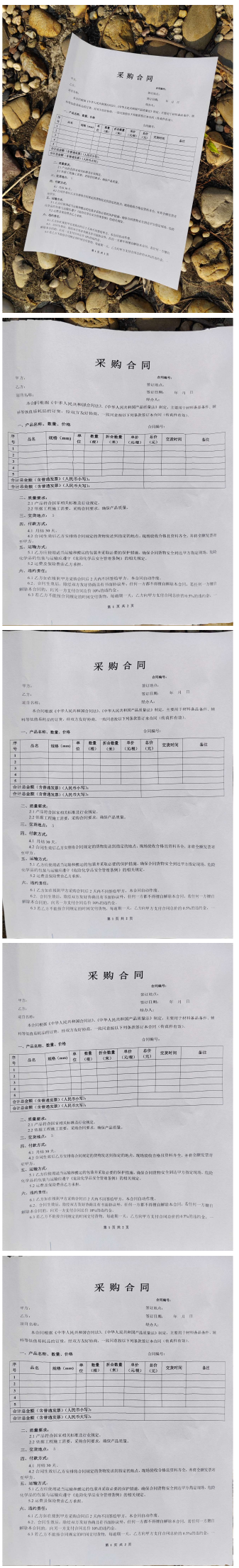}}
\subfloat{\includegraphics[scale=0.3]{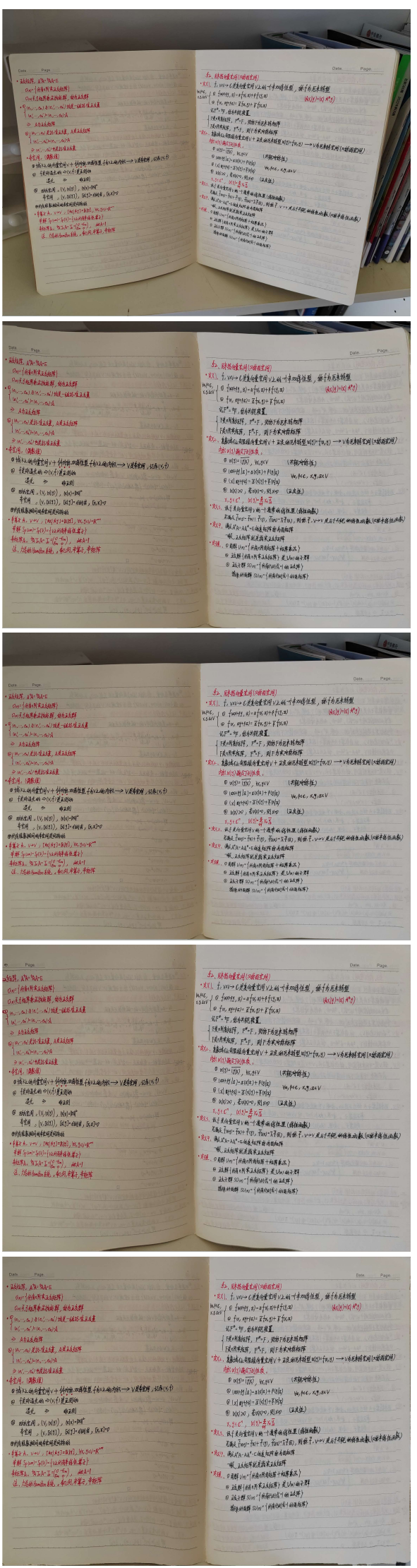}}
\subfloat{\includegraphics[scale=0.3]{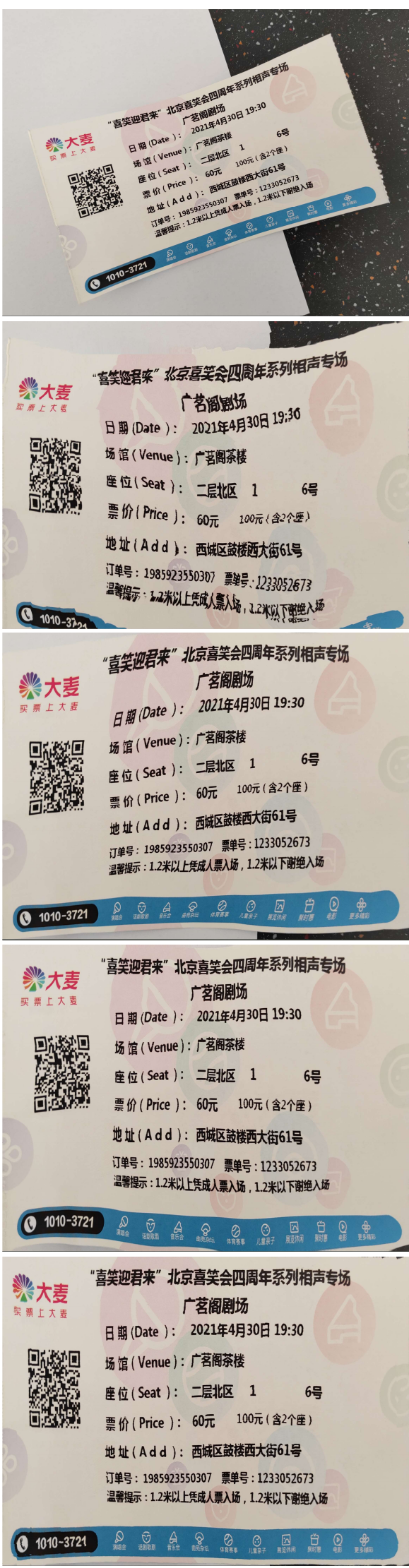}}
\caption{Qualitative visual comparison with existing methods on DocReal benchmark \cite{Yu_2024_WACV}.}
\label{DocReal_vis}
\end{figure}

\begin{figure}[t]
\centering
\subfloat{\includegraphics[width=0.13\linewidth]{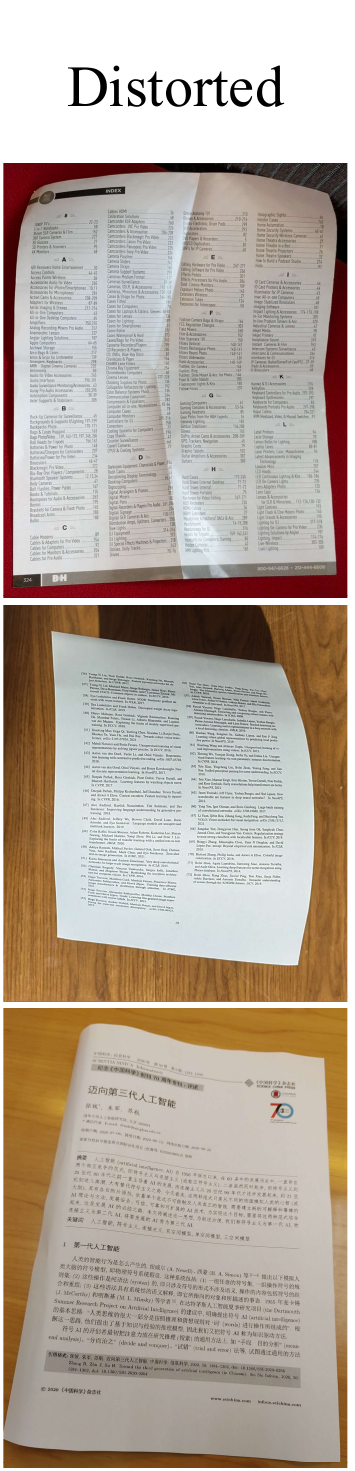}}
\subfloat{\includegraphics[width=0.13\linewidth]{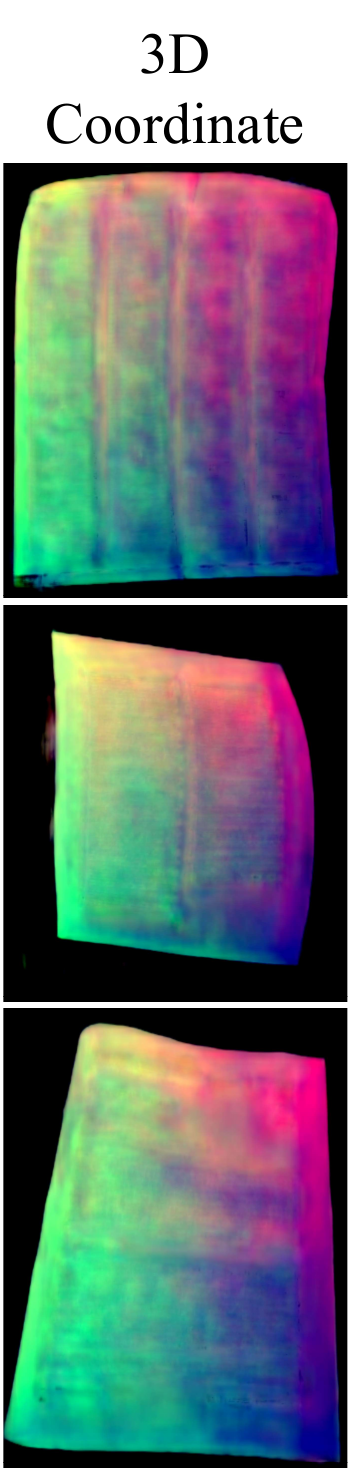}}
\subfloat{\includegraphics[width=0.13\linewidth]{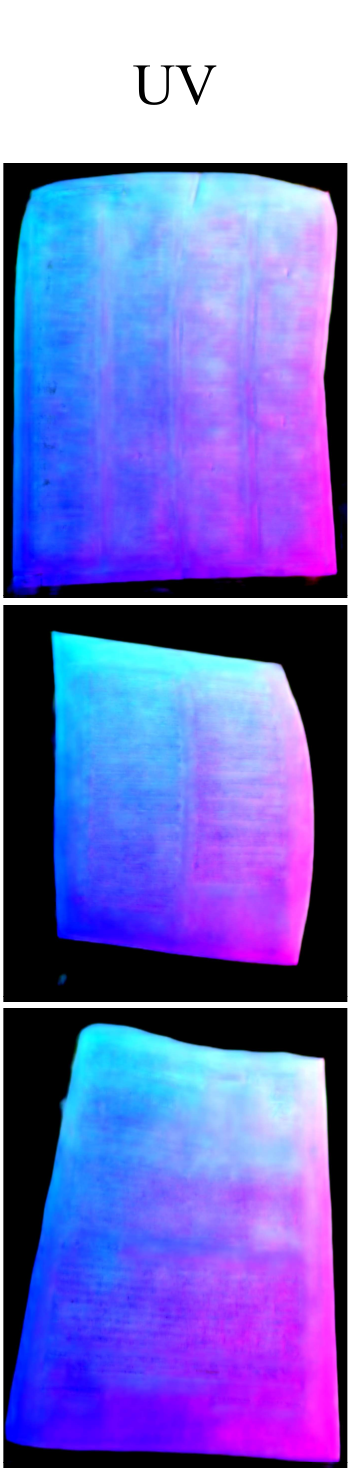}}
\subfloat{\includegraphics[width=0.13\linewidth]{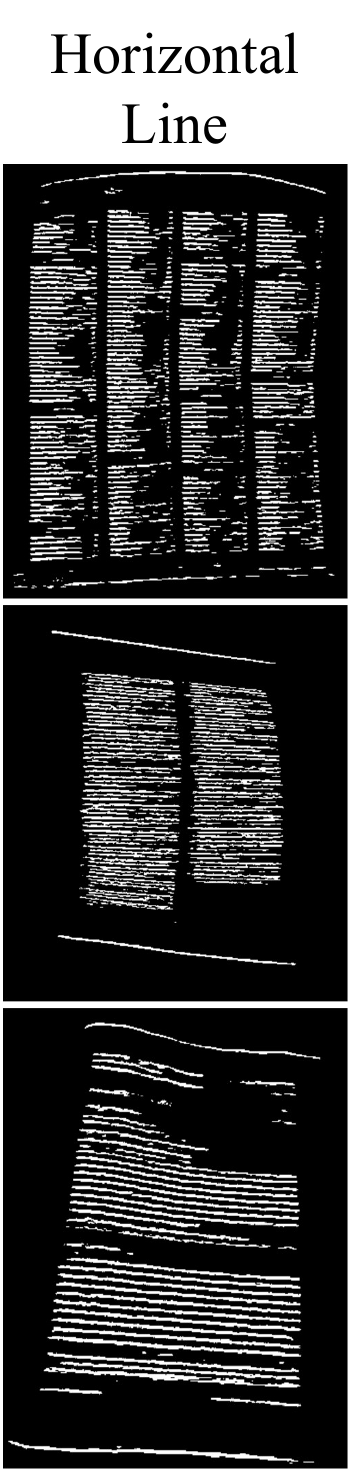}}
\subfloat{\includegraphics[width=0.13\linewidth]{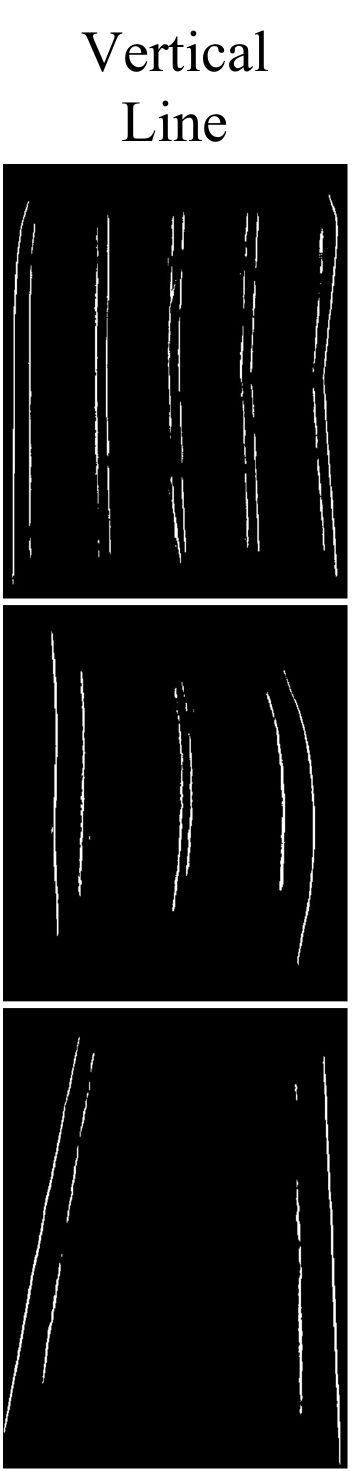}}
\subfloat{\includegraphics[width=0.13\linewidth]{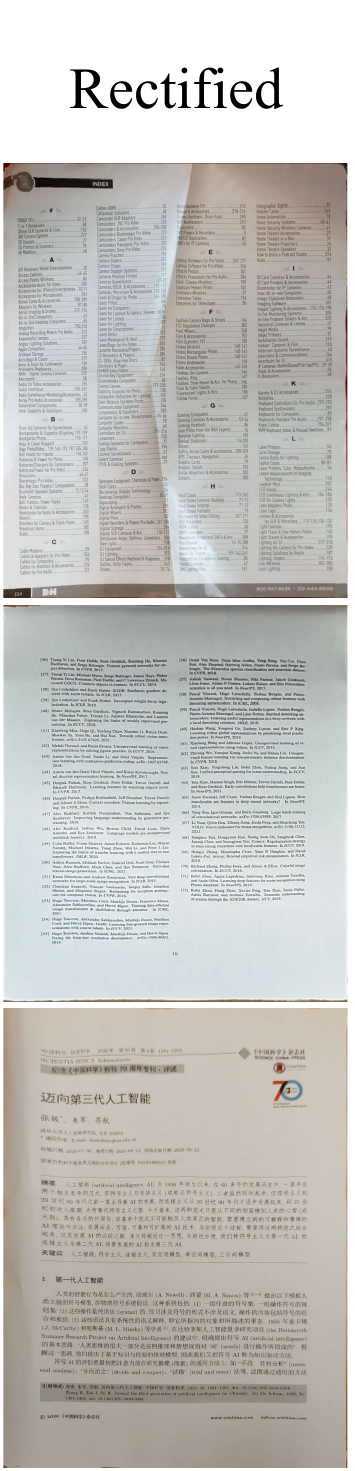}}
\caption{Predcit results of our proposed method. The first column ``Distorted'' is the given distorted document image and the last column ``Rectified'' represents the image rectified by our model. These three samples are from DocUNet \cite{Ma2018DocUNetDI}, DIR300 \cite{Feng2022GeometricRL} and DocReal \cite{Yu_2024_WACV} benchmark respectively.}
\label{pred_vis}
\end{figure}

\subsection{Experimental Results}
{\bf Performance on DIR300 Benchmark.} The results, detailed in Table \ref{tab:eval_dir300}, use different evaluation metrics where ``$\uparrow$'' indicates that higher values are better, and ``$\downarrow$'' indicates that lower values are better. Notable improvements are observed across all four evaluation metrics: MS-SSIM (+1.52\%), LD (-9.82\%), AD (-8.72\%), ED (-7.98\%), and CER (-20.77\%). Compared to FTDR \cite{li2023foreground}, which leverages cross-attention to capture document foreground and text line information, our model achieves a significant improvement in CER, outperforming it by 33.81\% (0.14 vs. 0.2115). 
For LA-DocFlatten \cite{li2023layout}, an approach designed to extract document layout features, our method enhances the perception of global and local features in terms of indicators measuring global similarity and local details (MS-SSIM/3.08\%, LD/9.82\%, AD/8.72\%). Qualitative visualization of the dewarping results are shown in Fig. \ref{DIR300_vis}, the first column is the input distorted document image, and the last column is the ground truth, also known as the flattened reference image.

These results demonstrate that our proposed method, leveraging inter-task feature aggregation, enhances the segmentation accuracy of document foreground and boundaries. 
In addition, our model architecture is robust to deformed images of different types and scenes, like folded and indoor (first two rows of Fig. \ref{DIR300_vis}), curved (last two rows), and outdoor grass (last row).

{\bf Performance on DocReal Benchmark.} Our method also demonstrates superior performance on this Chinese benchmark. Compared to the method proposed in the DocReal dataset \cite{Yu_2024_WACV}, which used 300K synthetic images for network training, our approach shows improvements of 5.36\%, 14.45\%, and 3.78\% in MS-SSIM, LD, and AD, respectively. These gains stem from the better local structure alignment of our rectified images with the reference images.
Similarly, when compared with RDGR \cite{Jiang2022RevisitingDI}, DocGeoNet \cite{Feng2022GeometricRL} and FTDR \cite{li2023foreground}---all of which employ text line attention---our method achieves state-of-the-art performance. 
The visual comparison is demonstrated in Fig. \ref{DocReal_vis}, showing the rectification results of our dewarped images with FTDR \cite{li2023foreground}, UVDoc \cite{UVDoc}, and DocRes \cite{Zhang2024DocResAG}. 
Our proposed model also exhibits superiority for images with complex ground background interference in natural scenes (first and second columns of Fig. \ref{DocReal_vis}), folded handwritten notes (third column), and tickets with affine transformation characteristics (last column).

{\bf Performance on DocUNet Benchmark.} We also perform quantitative comparisons with earlier methods on the DocUNet benchmark as shown in Table \ref{tab:eval_docunet}. For the line-focused methods RDGR \cite{Jiang2022RevisitingDI}, FTDR \cite{li2023foreground}, and DocGeoNet \cite{Feng2022GeometricRL}, we achieve at least 12.13\% and 12.89\% improvements in CER and ED on a 60-image OCR setting, respectively. For methods that introduce 3D coordinates or UV maps, such as DewarpNet \cite{Das2019DewarpNetSD}, Pice-Wise \cite{Das2021EndtoendPU}, LA-DocFlatten \cite{li2023layout}, and UVDoc \cite{UVDoc}, the rectification results of our framework also have a significant improvement in OCR performance (CER/13.33\%, ED/15.76\%).

{\bf Rectification and Intermediate Results.} We depict the prediction results of our proposed framework in Fig. \ref{pred_vis}. Each row represents an example from three different benchmarks. We show that our method can segment document boundaries clearly as well as local horizontal and vertical lines.

\begin{table*}[t]\rmfamily
\centering
\caption{Experiments on four tasks or different groups of 3D Coordinate (3D), UV map, horizontal and vertical lines (H-Line and V-line).  Experiment No. (\romannumeral 1) - (\romannumeral 4) are single auxiliary tasks. No. (\romannumeral 5) - (\romannumeral 6) are experiments grouped by global and local features. \textit{FA} and \textit{Gate} represent the Feature Aggregation module and Gating mechanism, respectively.} 
\label{tab:ablation_task}
\begin{tabular}{llcccccccccccc}
\toprule
\multirow{2}{*}{No.} & \multirow{2}{*}{Task(s)} & \multicolumn{3}{c}{DIR300 Benchmark}&\multicolumn{3}{c}{DocReal Benchmark}\\
& & MS-SSIM$\uparrow$ & LD$\downarrow$/AD$\downarrow$ & ED$\downarrow$/CER$\downarrow$ & MS-SSIM$\uparrow$ & LD$\downarrow$/AD$\downarrow$& ED$\downarrow$/CER$\downarrow$ \\
\midrule
\romannumeral 1 & \textit{3D} & 0.64 & 5.60/0.186 & 477.92/0.1654 & 0.56 & 9.25/0.239 & 202.68/0.2770 \\
\romannumeral 2 & \textit{UV} & 0.62 & 5.86/0.187 & 453.07/\underline{0.1556} & 0.56 & 9.03/0.242 & \textbf{192.06}/\underline{0.2623} \\
\romannumeral 3 & \textit{H-Line} & 0.63 & 5.69/0.182 & 494.24/0.2154 & 0.56 & 9.00/\underline{0.236} & 214.79/0.2931 \\
\romannumeral 4 & \textit{V-Line} & 0.63 & 5.60/0.183 & 580.48/0.1745 & 0.57 & 8.78/0.247 & 206.85/0.2816 \\
\midrule
\romannumeral 5 & \textit{3D \& UV} & \underline{0.65} & \underline{5.51}/0.187 & 501.21/0.2090 & 0.57 & 8.83/0.256 & 200.95/0.2805 \\
\romannumeral 6 & \textit{H-Line \& V-Line} & 0.62 & 6.11/0.184 & 473.40/0.1741 & 0.56 & 9.12/0.240 & 208.99/0.2808 \\
\romannumeral 7 & \textit{All four tasks} & 0.64 & 5.58/\underline{0.181} & \underline{451.96}/0.1646 & 0.57 & \underline{8.77}/0.241 & 
207.96/0.2887 \\
\midrule
\romannumeral 8 & \textit{All four tasks \& FA \& Gate} & \textbf{0.67} & \textbf{5.14}/\textbf{0.178} & \textbf{444.07}/\textbf{0.1400} & \textbf{0.59} & \textbf{8.41}/\textbf{0.229} & \underline{193.06}/\textbf{0.2604} \\
\bottomrule
\end{tabular}
\end{table*}

\subsection{Ablation Studies}
For these features closely related to distorted images, we first verify the role of each task and grouping according to global and local information on the correction task on the DIR300 \cite{Feng2022GeometricRL} and DocReal \cite{Yu_2024_WACV} benchmarks through ablation experiments. Then we perform ablation on the feature aggregation module (FA) and gating mechanism (Gate) we proposed.

{\bf Effect of Different Tasks or Groupings on the Rectification Results.} As shown in Table \ref{tab:ablation_task}, experiments (\romannumeral 1)-(\romannumeral 4) use 3D Coordinates, UV maps, horizontal lines (H-Line) and vertical lines (V-Line) as auxiliary information to guide the prediction of the two-dimensional deformation field. 
Experiments (\romannumeral 5) and (\romannumeral 6) group two global tasks and local tasks. All four tasks are predicted to regress the two-dimensional deformation field as shown in experiment (\romannumeral 7). 

From the experimental results, it can be found that for a single task, the UV map has a positive impact on the OCR performance (experiment (\romannumeral 2) in Table \ref{tab:ablation_task}). For the correction of local details, such as AD indicators, only using horizontal line information performs well (experiment (\romannumeral 3) in Table  \ref{tab:ablation_task}), reaching or approaching the best value on both benchamrks. 

For the combination of multiple tasks, 3D coordinates (3D) and UV map have a positive contribution to global correction (MS-SSIM in experiment (\romannumeral 5) in Table \ref{tab:ablation_task}). However, when 3D features are introduced on the UV map, experiment (\romannumeral 5), the performance of OCR will drop significantly. This shows that tasks are not always mutually beneficial, but have negative effects, and an effective fusion module is needed to optimize this phenomenon. In contrast, using only or in combination horizontal and vertical lines will have a slightly worse effect on the global perception of distorted documents (experiments (\romannumeral 3), (\romannumeral 4) and (\romannumeral 6) in Table \ref{tab:ablation_task}).

In addition, it is worth noting that for experiment (\romannumeral 7) in Table \ref{tab:ablation_task}, when all four features are used to guide the anti-distortion task, the accuracy of OCR does not reach the best. Therefore, simply concatenating multiple features cannot be expected to improve the performance of the distorted document image unwarping task. The feature aggregation module and gating mechanism proposed in our architecture are based on these findings (experiment (\romannumeral 8) in Table \ref{tab:ablation_task}). Under the guidance of multi-task learning, these two modules will minimize the interference between auxiliary information and extract the most useful features for the final task.

\begin{table*}[t]\rmfamily
\centering
\caption{Ablation of feature aggregation (FA) module and gating mechanism (Gate) on DocUNet \cite{Ma2018DocUNetDI}, DIR300 \cite{Feng2022GeometricRL} and DocReal \cite{Yu_2024_WACV} Benchmarks. ``\ding{51}'' indicates that the module of FA or Gate is used. ``\ding{55}'' is the opposite. ``\#Param'' denotes the number of model parameters.}
\label{ablation_fa_gate}
\begin{tabular}{cccccccccccccc}
\toprule
\multirow{2}{*}{\textit{FA}} &\multirow{2}{*}{\textit{Gate}} & \multirow{2}{*}{\textit{\#Param}} & \multicolumn{3}{c}{DocUNet Benchmark}&\multicolumn{3}{c}{DIR300 Benchmark}&\multicolumn{3}{c}{DocReal Benchmark} \\

& & & MS$\uparrow$ & LD/AD$\downarrow$ & ED/CER$\downarrow$ & MS$\uparrow$ & LD/AD$\downarrow$ & ED/CER$\downarrow$ & MS$\uparrow$ & LD/AD$\downarrow$& ED/CER$\downarrow$ \\
\midrule

\ding{55} & \ding{55} & 21.1M &  0.50 & 7.28/0.33 & 350.5/0.14 & 0.64 & 5.58/0.18 & 452.0/0.17 & 0.57 & 8.77/0.24 & 208.0/0.29 \\
\ding{51} & \ding{55} & 22.7M & 0.50 & 7.30/0.32 & 376.8/0.14 & 0.64 & 5.56/0.18 & 435.8/0.17 & 0.58 & 8.67/0.23 & 201.6/0.28 \\

\ding{55} & \ding{51} & 28.2M &  0.49 & 7.49/0.32 & 384.8/0.18 & 0.64 & 5.56/0.18 & \textbf{399.6}/0.15 & 0.57 & 8.90/0.24 & 209.0/0.28 \\

\ding{51} & \ding{51} & 29.8M &  \textbf{0.51} & \textbf{7.10}/\textbf{0.31} & \textbf{330.2}/\textbf{0.13} & \textbf{0.67} & \textbf{5.14}/\textbf{0.18} & 444.1/\textbf{0.14} & \textbf{0.59} & \textbf{8.41}/\textbf{0.23} & \textbf{193.1}/\textbf{0.26}
 \\
\bottomrule
\end{tabular}
\end{table*}

{\bf Ablation of Feature Aggregation (FA) Module and Gate.} To demonstrate the effectiveness of our proposed method, we selectively use or do not use the feature aggregation module (FA, 1.6M parameters) or the gating mechanism (Gate, 7.1M parameters) under the premise of predicting four tasks. The experimental results are shown in Table \ref{ablation_fa_gate}. For these three benchmarks, when only one of the modules is merged into the entire pipeline, although it has a certain effect on several evaluation metrics, it is still not significant enough. In general, the two modules jointly serve the dewarping task, and almost all indicators have obvious improvements. 
Compared with directly predicting the 2D deformation field without feature fusion (first row in Table \ref{ablation_fa_gate}), the MS-SSIM on the three benchmarks increased by 3.40\% on average.
LD and AD also achieved 5.05\% and 4.42\% improvement respectively.
In terms of text recognition accuracy, the Edit Distance (ED) achieved a better performance of 4.9\% evenly. The Character Error Rate (CER) reached an impressive 10.3\%.






\begin{figure}[!t]
\centering
\subfloat{\includegraphics[width=0.167\linewidth]{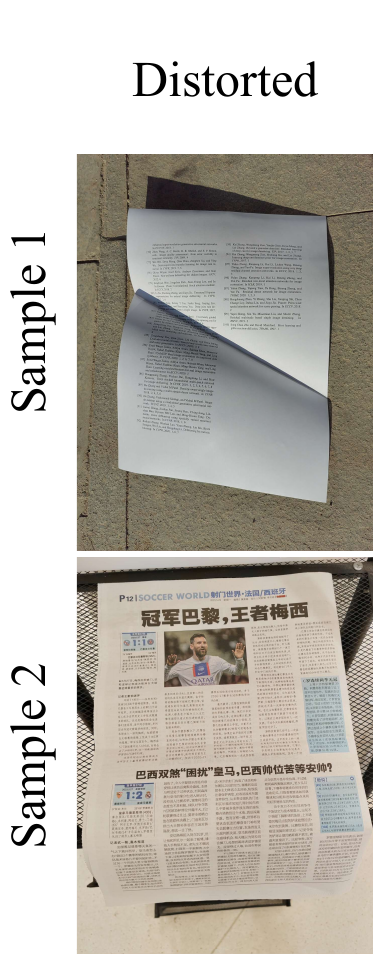}}
\subfloat{\includegraphics[width=0.135\linewidth]{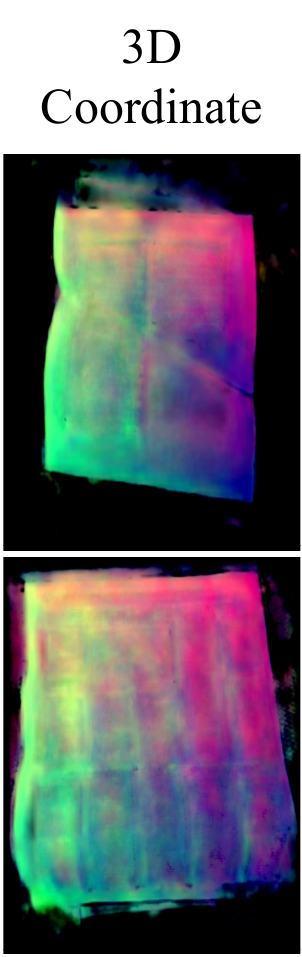}}
\subfloat{\includegraphics[width=0.135\linewidth]{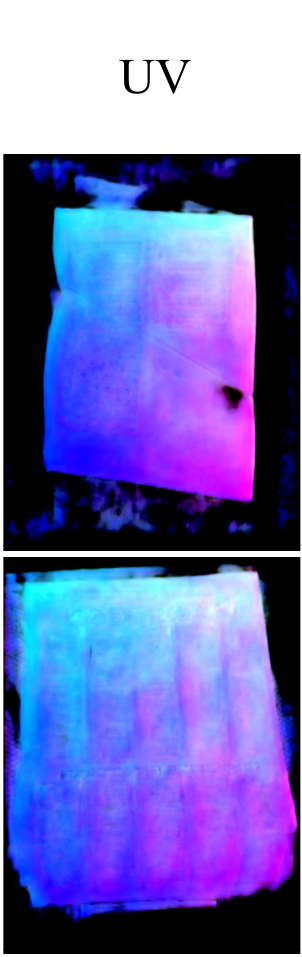}}
\subfloat{\includegraphics[width=0.135\linewidth]{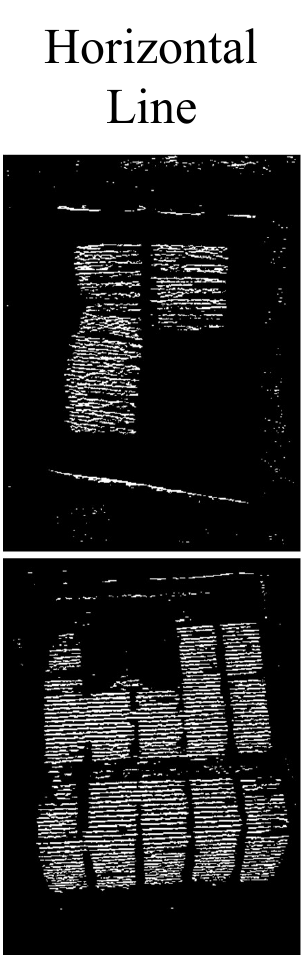}}
\subfloat{\includegraphics[width=0.135\linewidth]{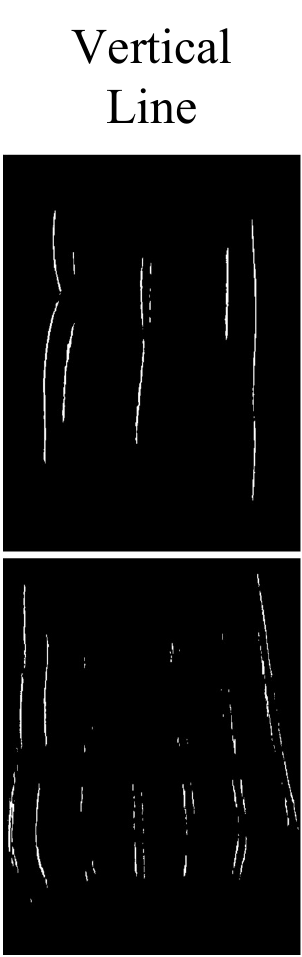}}
\subfloat{\includegraphics[width=0.135\linewidth]{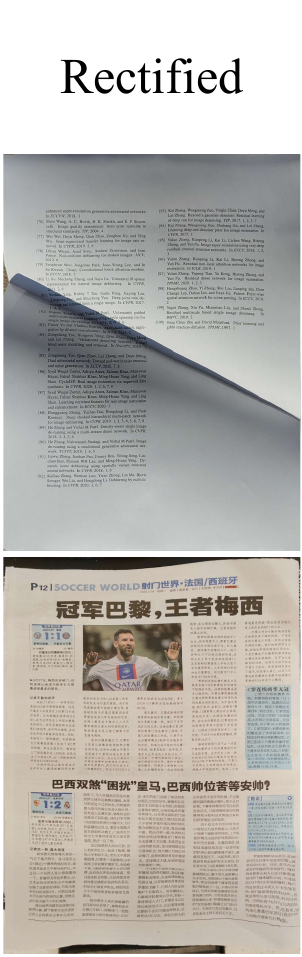}}
\caption{A few images are used for sample analysis. Sample 1 and 2 are sampled from DIR300 \cite{Feng2022GeometricRL} and DocReal \cite{Yu_2024_WACV} benchmark. \textit{Sample 1} image (first row) has strong folding except for shadows. \textit{Sample 2} (second row) has grid-shaped texture interference in the background, and the background color is close to the edge color of the foreground paper.}
\label{bad_samples_vis}
\end{figure}

\subsection{Limitations} Through the above experiments and ablation analysis, although our proposed method has achieved certain advanced performance in the task of distorted document image correction compared with previous methods. But there are still some limitations. We analyze several examples that are not well rectified as detailed in Fig. \ref{bad_samples_vis}.

\textit{Sample 1} in the first row of the figure is affected by lighting conditions and partial shadows, causing disturbt in the 3D coordinates and UV map outputs. 
However, the prediction of horizontal and vertical lines can adjust this inaccuracy, thereby‌ the dewarped image can still be clear and accurate to the edge of the document foreground. 
Besides, the paper in this sample is also severely folded. The text that is blocked by folding cannot be corrected well.

For \textit{Sample 2} in Fig. \ref{bad_samples_vis}, it is a newspaper with dense text content. The background has a grid-shaped interference, and at the lower boundary of the document, the background color is very close to the color of the paper. Despite our method being able to accurately detect the text lines and a few vertical lines inside the document, the upper and lower boundaries of the foreground are still not clearly segmented.

\section{Conclusion}
We proposed SalmRec, an adaptive task-selective distorted document correction architecture based on end-to-end multi-task learning. Our method can effectively and automatically refines the aggregation of multiple features across different granularities. The proposed gating mechanism is used to control the output of information of similar granularity. Extensive experiments on multiple benchmarks show that our network is capable of aggregating the respective advantages of multiple tasks and alleviate the mutual interference within the tasks. 
For future work, it is possible to explore a more lightweight yet robust network for correcting distorted document images with complex backgrounds.

\bibliographystyle{cas-model2-names}

\bibliography{cas-refs}


\end{sloppypar}
\end{document}